\PassOptionsToPackage{usenames,table}{xcolor}
\documentclass[xcolor]{bmvc2k}
\usepackage[usenames,table]{xcolor}
\usepackage{placeins}
\usepackage{graphicx}
\usepackage{amsmath,amssymb,amsthm,amscd}
\usepackage{gensymb}
\usepackage{rotating}
\usepackage{multirow}
\usepackage{multicol}
\usepackage{textcomp}
\usepackage{cleveref}


\newcommand{\off}[1]{}

\newcommand{\region}{\mathcal{R}}
\newcommand{\qegion}{\mathcal{Q}}

\newcommand{\latinphrase}[1]{\emph{#1}}

\newcommand*{\etc}{\latinphrase{etc.}\@\xspace}

\definecolor{attr}{rgb}{0.85,0.88,0.90}
\definecolor{code}{rgb}{0.95,0.93,0.90}

\newcommand{\detnet}{\textsc{DNet}\xspace}

\newcommand{\dataset}{\textsc{HPatches}\xspace}
\newcommand{\datasetdet}{{HPSequences}\xspace}

\newcommand{\cittilde}{\scriptsize\citep{Verdie15}}
\newcommand{\citfast}{{\scriptsize\citep{rosten10faster}}}
\newcommand{\citsurf}{\scriptsize\citep{bay2006surf}}

\newcommand{\citbrisk}{\scriptsize\citep{BRISK}}
\newcommand{\cithessian}{\scriptsize\citep{mikolajczyk02affine}}
\newcommand{\citdog}{\scriptsize\citep{lowe04distinctive}}
\newcommand{\cittcdet}{\scriptsize\citep{zhang2017learning}}
\newcommand{\citlift}{\scriptsize\citep{yi16lift}}
\newcommand{\citdepdet}{\scriptsize\citep{Lenc16}}

\def\eg{\emph{e.g}\bmvaOneDot}

\renewcommand{\paragraph}[1]{\par\smallskip\noindent\textbf{#1}}

\title{Large scale evaluation of local image feature detectors on homography datasets}
\addauthor{Karel Lenc}{http://www.robots.ox.ac.uk/~karel/}{1}
\addauthor{Andrea Vedaldi}{http://www.robots.ox.ac.uk/~vedaldi/}{1}
\addinstitution{
 Department of Engineering Science\\
 University of Oxford\\
 Oxford, UK
}
\runninghead{Lenc, Vedaldi}{Large scale evaluation of local feature detectors}

\begin{document}
\maketitle
\begin{abstract}
We present a large scale benchmark for the evaluation of local feature detectors. Our key innovation is the introduction of a new evaluation protocol which extends and improves the standard detection repeatability measure. The new protocol is better for assessment on a large number of images and reduces the dependency of the results on unwanted distractors such as the number of detected features and the feature magnification factor. Additionally, our protocol provides a comprehensive assessment of the expected performance of detectors under several practical scenarios. Using images from the recently-introduced \emph{HPatches} dataset, we evaluate a range of state-of-the-art local feature detectors on two main tasks: viewpoint and illumination invariant detection. Contrary to previous detector evaluations, our study contains an order of magnitude more image sequences, resulting in a quantitative evaluation significantly more robust to over-fitting. We also show that traditional detectors are still very competitive when compared to recent deep-learning alternatives.
\end{abstract}

\section{Introduction}\label{sec:intro}

Despite advances in distributed representations such as deep convolutional networks, local viewpoint invariant features still play an important role in tasks such as structure from motion and image retrieval. In these applications, deep learning has often been used to \emph{improve} rather than to \emph{replace} local features. While most of this work focused on learning feature \emph{descriptors}, recently there has been progress in learning \emph{detectors} as well. For example, in~\cite{Verdie15} use deep networks to learn a local feature detector robust to illumination changes, \cite{Yi16Learning} for orientation assignment, \cite{Lenc16} for learning detectors without supervision, and \cite{yi16lift} for learning local feature detectors, orientation assignment and descriptors.

An obstacle to further progress in learning local feature detectors is the lack of a modern, large-scale evaluation benchmark for this task. Advances in tasks such as image classification were driven by the introduction of benchmarks such as ImageNet. For local feature descriptors, recent contributions such as~\dataset~\cite{Balntas17} may play a similar role, but there is still no good solution for detection. Several works for testing performance of both detector and descriptors emerged, \cite{schoenberger2017comparative, mishkin2015wxbs}, however we believe that being able to test and compare algorithms separately provides invaluable insight into where the progress is made.

In order to address this shortcoming, in this paper we propose \textbf{a modern evaluation of feature detectors}. We do so by augmenting the evaluation protocol (\cref{sec:methods}) of feature benchmarks which come with ground truth homographies for image sequences representative of various difficult imaging scenarios, such as illumination and viewpoint changes. We build especially on the recently-introduced~\dataset dataset; however, while the latter  contains pre-detected image patches for descriptor evaluation, we discard such patches and use the images as a whole to assess feature detectors instead. We further refer to this dataset as \datasetdet.

For the evaluation protocol, we start from the detector repeatability evaluation protocol introduced in the classic paper of~\cite{Mikolajczyk05}, as it is an accepted standard, and we improve it in various ways. Specifically, compared to earlier benchmarks such as VGG Affine, which are nowadays heavily over-fitted due to their small size and due to having been used by the community for many years, \datasetdet is much larger and less prone to over-fitting. Furthermore, we improve the evaluation protocol by addressing issues in the invariance to feature magnification factor found  in the reference implementation of repeatability (\cref{s:det_magfac}). We also propose to modify the protocol to explicitly control for the number of detected features per image (\cref{s:det_thr}), yielding fairer detector comparisons. Due to the significantly increased number of images compared to VGG Affine (696 vs 48), we also change the way results are aggregated, reported, and analysed, comparing detectors quantitatively using a single plot (\cref{s:det_analysis}). Additionally, we include trivial baselines based on random features which provide lower bounds of the expected performance. 
The new benchmarking code for automatic evaluation of detectors is released in the open source domain, simplifying reproducibility of the future research.
We aim to provide a robust, easy-to-reproduce and easy-to-use evaluation platform for comparison of local feature detector performance on planar scenes. Both source code and pre-computed scores used for this manuscript are freely available\footnote{\url{https://github.com/lenck/vlb-deteval}}.
 %results are computed using the open-source framework VLB~\footnote{\url{https://github.com/lenck/vlb}}.

Having designed a suitable benchmark, our second contribution is to \textbf{analyse} classic feature detector against modern ones based on \textbf{deep learning} (\cref{sec:exps}). We find that learning detectors significantly improves robustness to illumination changes, but that, for viewpoint invariance, traditional detectors using scale selection and affine adaptation are still nearly as good and sometimes better than learned ones.

\section{Related work}\label{sec:related}
In this section we introduce evaluated local feature detectors (\cref{sec:r_dets}) and existing benchmarks for their evaluation (\cref{sec:r_benchmarks}). 

\subsection{Local detectors}\label{sec:r_dets}
Local image feature detectors differ by the type of features that they extract, \eg points~\citep{harris88combined,rosten10faster}, circles~\citep{lowe04distinctive,mikolajczyk01harris-laplace}, or ellipses~\citep{baumberg00reliable,matas02local,mikolajczyk02affine}. In turn, the type of feature determines which class of transformations that they can handle: Euclidean transformations, similarities, and affinities respectively. Additionally, we can divide detectors as follows:

\paragraph{Hand-crafted detectors.} Standard, hand-crafted local feature detectors vary based on the visual structures used as anchors for the features, \eg corners or other operators of the image intensity such as the \emph{Hessian of Gaussian}~\citep{beaudet78rotationally} or the \emph{structure tensor}~\citep{forstner86a-feature,harris88combined,zuliani05mathematical}. Going beyond roto/translation, scale selection methods using the \emph{Laplacian/Difference of Gaussian} operator (L/DoG) or Hessian of Gaussian were introduced in~\citep{lowe04distinctive,mikolajczyk01harris-laplace} and further extended with \emph{affine adaptation}~\citep{baumberg00reliable,mikolajczyk02affine} to handle full affine transformations. 

\paragraph{Accelerated detectors.} Machine learning can be used to imitate and accelerate an off-the-shelf detector defined \emph{a-priori}~\citep{dias95a-neural,rosten06machine,sochman09learning,holzer12learning}. \citet{rosten10faster} use simulated annealing to optimise the parameters of their FAST detector for repeatability. For the SURF detector \citep{bay2006surf}, the authors use integral images to approximate the Hessian feature response.

\paragraph{Learned detectors.} Learning detectors attempts to discover or improve the visual anchors used for detection, a task much harder than using hand-crafted anchors. Early attempts used genetic programming~\citep{trujillo06synthesis,olague11evolutionary-computer-assisted}. More recently, \citet{Yi16Learning} learn to estimate the orientation of feature points using deep learning. A related approach is the TILDE detector~\citep{Verdie15} for illumination invariance. The LIFT framework~\citep{yi16lift} aims at learning detector, descriptor and orientation estimation jointly using patches, while SuperPoint \cite{DeTone_2018_CVPR_Workshops} uses full images.
Another approach to unsupervised learning of keypoint detectors is \detnet \cite{Lenc16}, which is trained using the covariance constraint and no supervision. A version of this detector is TCDET~\citep{zhang2017learning}, combined geometry and appearance losses. The covariant constraint is extended for affine adaptation in \cite{mishkin2017learning}.

\subsection{Evaluation of local detectors}\label{sec:r_benchmarks}
The standard protocols for the evaluation of local feature detectors and descriptors was established by \cite{Mikolajczyk05, km2005pami} using the VGG Affine dataset, which contains 8 sequences of 6 images related by a known homography.
% The advantage of this approach is that defines a pixel-to-pixel mapping, a property which is not possible to obtain for different 3D scene models (such as Epipolar geometry) and allows measuring repeatability between two images from a sequence.
Detectors are assessed in terms of their \emph{repeatability}, which measures their robustness to nuisance effects such as a change in viewpoint or illumination. The standard definition of repeatability has some shortcomings. First, features are compared by the overlap of their support, generally elliptical, which may not encode all relevant geometric information (e.g.\ it disregards the feature orientation) and depends on the size of regions, which is arbitrary and requires normalisation. Second, computing repeatability is somewhat slow and uses in practice a number of approximations, which we show in this paper are not innocuous.

Many datasets followed the introduction of VGG Affine.
In the \emph{Hanover dataset} \citep{cordes2013high}, the number of sequences is extended while improving the precision of the homography.
While the traditional and most commonly used VGG Affine dataset contains images that are all captured by a camera,  the \emph{Generated Matching dataset}~\citep{Fischer2014} is obtained by generating images using synthetic transformations.
The \emph{Edge Foci dataset}~\citep{zitnick2011edge} consists of sequences with very strong changes in viewing conditions, making the evaluation somewhat specialized to extreme cases; furthermore, the ground truth for  non-planar scenes does not uniquely identify the correspondences since the transformations cannot be well approximated by homographies.
In the \emph{Webcam dataset}~\citep{Verdie15}, new sequences for testing illumination changes are presented.
The \emph{DTU robots dataset}~\citep{aanasinteresting} goes beyond homographies and uses scenes with a known 3D model, obtained using structured lighting.
In \cite{mishkin2015wxbs}, the authors introduce a new dataset for generalised wide baseline stereo matching across geometry (homography and epipolar), illumination, appearance over time and capturing modes. 

Instead of introducing new evaluation protocols, our main goal is to provide a large scale evaluation baselines over multiple datasets. Additionally we improve the repeatability score and quantitatively analyse results across different tasks.

%\todo{Explain better how our dataset differs <- it's mostly the protocol that differs, dataset is just bigger}

%In this section we perform evaluation on the \datasetdet dataset. Compared to the datasets from the previous section, this dataset provides more test sequences (\ie a larger variety of tested scenes) and has a strict division between illumination and viewpoint sequences

\section{Evaluation protocol: repeatability revisited}\label{sec:methods}

% TODO: See https://arxiv.org/pdf/1504.07967.pdf

In this section we refresh the traditional repeatability evaluation method proposed by~\citet{Mikolajczyk05}, addressing some of its shortcomings and improving its applicability to large datasets.

Given an image $I$, a detector extracts a set $\mathcal{D}=\{\region_1,\dots,\region_n\}$ of $n$ regions $\region_i \subset \mathbb{R}^2$, generally ellipses. Given a second image $I'$ related to the first by an homography, the same detector extracts another list $\mathcal{D'}=\{\region'_1,\dots,\region'_n\}$ of $m$ regions. Following~\citet{Mikolajczyk05}, the repeatability score $\mathit{rep}(\mathcal{D},\mathcal{D'},H)$ for the detected features is the fraction of features that match between images with sufficient geometric overlap up to the homography $H$. While the concept is simple, there are many important implementation details that strongly affect the outcome.\footnote{The actual implementation slightly differs from the definition in the paper~\citet{Mikolajczyk05} which also lacks some details; our description follows the authoritative implementation by the same authors.} These details are discussed next.

% TODO: 30 pixel radius or rectangle? The code may differ from the paper
The degree of geometric match between two regions $\region,\qegion\subset\mathbb{R}^2$ is given by their \emph{overlap}  $o(\region,\qegion) = |\region\cap \qegion|/|\region\cup \qegion|$. If $H$ is the homography transformation that reprojects pixels from image $I'$ back to image $I$, the overlap measure can be changed to $o(\region,H\region')$ to compensate for this transformation. However, as noted by \citet{Mikolajczyk05}, overlap can generally be increased just by scaling (magnifying) the detected features by a constant \emph{magnification factor} $s\in\mathbb{R}_+$, which can be trivially incorporated in the definition of any detector. For example, if $s\region$ denotes the effect of scaling the region $\region$ by a factor $s$ around its center of mass, and if $\region$ and $\region'$ differ only by a shift, then $\lim_{s\rightarrow\infty}o(s\region,Hs\region')=1$. \citet{Mikolajczyk05} address this issue by rescaling features so that the first one has an area of $30^2$, resulting in the (asymmetric) normalised overlap score $o(\region,\region'|H) = o(s(\region)\region,Hs(\region)\region')$, where $s(\region) = 30^2/|\region|$. Please note that this does not fully remove the influence of the detected scale, as the relative scale between the compared regions is still important (as the normalisation constant for both regions is $s(\region)$).

Next, in order to compute repeatability in two feature sets, features must be matched based on ellipse overlap \footnote{Several works \cite{rosten10faster, Ono2018} use only distance of the keypoint centres, however it is only applicable for joint detector and descriptor evaluation.}.
In order to do so, features that do not belong to the common part of $I$ and $I'$ are dropped as they cannot be matched. This is done by sending the center of each region $\region'$ to $I$ using $H$ and testing for inclusion in the domain of $I$; the same operation is repeated for regions $\region$ in the other direction. Let $\mathcal{D}_c$ and $\mathcal{D}'_c$ be the remaining features. Pairs of such regions are associated with score $s(\region_i,\region_j') = o(\region_i,\region'_j|H)$ if their normalised overlap is at least $1-\epsilon_O$ and $s(\region_i,\region_j') = -\infty$ otherwise. The matches $\mathcal{M}^* \subset \mathcal{D}_c\times\mathcal{D}'_c$ are determined as the bipartite graph that maximises\footnote{In practice, bipartite matching is approximated greedily.} the overall score $\sum_{(\region,\region')\in\mathcal{M}} s(\region,\region')$. Note that this maximization retains only pairs with overlap above the threshold and matches each region at most once. Finally, \emph{repeatability} is defined as $\mathit{rep}(\mathcal{D},\mathcal{D'},H)=|\mathcal{M}^*|/\min\{|\mathcal{D}_c|,|\mathcal{D}_c'|\}$.

\subsection{Magnification factor invariance}\label{s:det_magfac}

\begin{figure}
	\centering
	\includegraphics[width=0.9\linewidth]{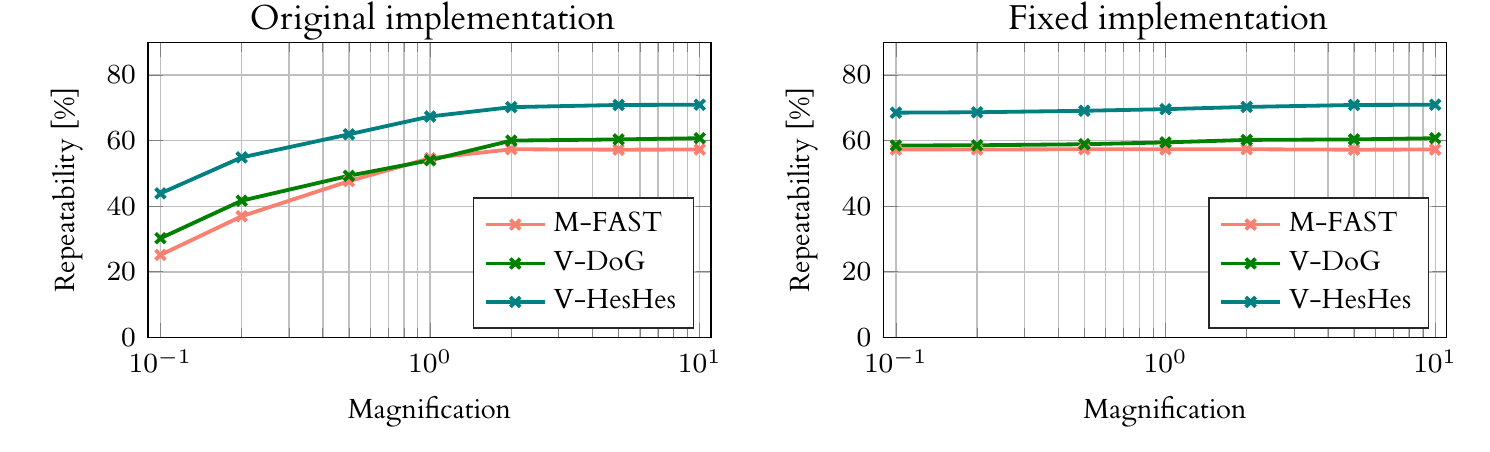}
	\vspace{-1.em}
	\caption{Average detector repeatability on VGG Affine for three detectors and increasing magnification factor (log scale). Due to normalisation, the lines should be approximately constant (right) but this is not the case in the original implementation (left) due to approximations.}\label{fig:rep-magnif}
\end{figure}

While in principle the use of a normalised overlap measure should make repeatability invariant to the detector magnification factor, the reference implementation of this measure still has a strong empirical dependency on this parameter, as can be seen in \cref{fig:rep-magnif}-left. We have identified that the cause of this issue is in the heuristic used for accelerating the ellipse overlap computation. This heuristic filters out ellipse pairs whose enclosing circles cannot overlap, a test which can be calculated quickly. However, in the original implementation of the test, this heuristic was applied \emph{before} the ellipse normalisation step. This leads to ellipses with area smaller than $30^2$ pixels being mistakenly skipped as unable to overlap, reducing the repeatability score. After fixing this relatively minor issue (by normalising the excircles), in \cref{fig:rep-magnif}-right we show that the repeatability becomes invariant to the magnification factor.

\subsection{Detection threshold}\label{s:det_thr}

Many local feature detectors have a single parameter that controls selectivity, which we generically call \emph{detection threshold} $\tau$. In hand-crafted detectors, $\tau$ is usually the minimum value of the cornerness measure, such as DoG, Hessian or structure tensor \etc, for which a feature is retained.

One would expect a detector to provide stable performance across all its detection thresholds. In practice, however, this might not be reflected by repeatability. In fact, with an increased number of features, it becomes easier to match features by accident, making repeatability biased for settings that produce more features. That is why, for a fair comparison, local feature detectors need to return a similar number of features and we test random detectors to obtain a baseline performance.

Since each detection algorithm anchors features to different visual primitives in an image, the number of detected features cannot be equalized by choosing a constant $\tau$ per detector for the whole dataset. Instead, similarly to~\citep{Verdie15, mishkin2015wxbs}, we run each detector to extract as many features as possible (by lowering $\tau$) and then consider only the top-$n$ detections from each image ranked by detection score, where $n \in \mathcal{N}=\{100, 200, 500, 1000\}$.\footnote{The upper limit 1000 was selected empirically, as some detectors produce fewer features even at the lowest $\tau$ than others.} Testing different values of $n$ is useful because the number of detections per image may differ based on the application and shows whether the detection score is predictive of the detected regions repeatability. As far as we are aware, testing the detector performance over various operational point is not a standard practice in local feature evaluation.

\subsection{Aggregated metrics and their analysis}\label{s:det_analysis}

\begin{table}
	\centering
	\begin{minipage}{0.5\linewidth}
		\footnotesize
	\begin{tabular}{|l | r r r|} \hline
		Dataset & \# Seq. & \#Ims. & \#Im. pairs \\\hline
		VGG Affine \cite{Mikolajczyk05} & 8 & 48 & 40 \\
		Webcam \cite{Verdie15} & 6 & 250 & 125 \\
		\datasetdet \cite{balntas2016learning} & 116 & 696 & 580 \\\hline
	\end{tabular}
	\end{minipage}
\begin{minipage}{0.4\linewidth}
	\caption{Basic statistics of the selected datasets for local feature detector evaluation.}\label{tab:datasets-det}	
\end{minipage}
\end{table}

So far, we have explained how to compute repeatability for a pair of images. Here we look at how a large dataset of image pairs can be used for assessment.

The benchmark of~\citet{Mikolajczyk05} contains a number of \emph{image sequences} $(I_t)_{t=0}^T$ and their evaluation reports repeatability for each sequence, fixing $I_0$ as reference image and varying $I_t,t=1,\dots,T$. Each sequence tests a particular aspect of feature detection, such as invariance to viewpoint, illumination, or noise changes. Furthermore, images in each sequence are sorted by the size of the nuisance variation, so plotting repeatability against $t$ normally shows a progressive reduction in repeatability.

Such an approach is suitable for VGG Affine, which contains just 8 sequences with 6 images each. Clearly, however, it does not scale well to larger datasets. Furthermore, it does not provide a single performance metric per detector, nor corresponding confidence margins, which makes it difficult to compare detectors' performance and to know how significant the differences are. Another  issue is that in datasets such as Webcam~\citep{Verdie15} and~\datasetdet~\cite{balntas2016learning} images cannot be easily sorted by the size of the nuisance variation, thus plotting repeatability against $t$ is meaningless.

We approach this issue by computing aggregated statistics over multiple images and factors of variation, similar to~\citep{Yi16Learning,zitnick14edge,Verdie15}. As repeatability is very sensitive to the number of features extracted, we also compute an average over different detection thresholds and analyse the distributions of repeatability scores so obtained to extract confidence margins.

In more detail, we are given a dataset which consists of a set of image pairs and homographies $\mathcal{T} = \{(I_1, J_1 | H_1), \dots, (I_T, J_T | H_T)\}$ (\cref{tab:datasets-det}). We will denote $\mathit{rep}(d,t,n)$ as the repeatability of a detector $d$, task $t \in \mathcal{T}$ and number of detections per image $n \in \mathcal{N}$. To score a detector $s$, we average repeatability across tasks and number of detections:
\begin{equation}\label{e:rep}
\mathit{rep}(d,n) =  |\mathcal{T}|^{-1} \sum_{t\in\mathcal{T}}{\mathit{rep}(d,t,n)},
\quad
\mathit{rep}(d) =  |\mathcal{N}|^{-1} \sum_{n\in\mathcal{N}}{\mathit{rep}(d,n)}.
\end{equation}
An ideal detector in our evaluation has a high average repeatability. Additionally, we consider also the \emph{variance} of the repeatability score, as a low variance means that the detector performance is consistent across different cases. We visualise both average and variance using box-and-whisker diagrams, plotting repeatability on the $x$ axis and detectors on the $y$ axis (\cref{fig:rep-hpatches}). These diagrams summarise at a glance the statistics $\mathit{rep}(d, t, n)$ for each detector $d$. The box percentiles are $25\%$ and $75\%$ (first and third quartile) and the whisker percentiles are $10\%$ and $90\%$. The length of the whiskers correspond to the length of the distribution tail. Additionally, we show the median (solid line) and the mean (red cross) of each distribution. We vary the line style of the whiskers to group detectors by type, generally based on their purported invariance (dotted for translation, dash-dot for scale, and dashed for affine invariant detectors).
Finally, for each detector, we show $\mathit{rep}(d,n)$ using box markers: {\footnotesize$\boxed{\mathrm{.1k}}$}  for $\mathit{rep}(d,100)$,  {\footnotesize$\boxed{\mathrm{.5k}}$} for $\mathit{rep}(d,500)$, and {\footnotesize$\boxed{\mathrm{1k}}$} for $\mathit{rep}(d,1000)$.

\paragraph{Stability error across detection thresholds.} To quantify the stability of the detector performance across detection thresholds, we calculate the detector instability as the standard deviation of the detector repeatability across different numbers of features, normalised by the average repeatability:
\begin{equation}\label{e:stb}
\mathit{stb}(d) = \mathit{rep}(d)^{-1} \cdot \sqrt{|\mathcal{M}|^{-1} {\textstyle\sum}_{n}{ \left[\mathit{rep}(d,n)-\mathit{rep}(d)\right]^2}}.  
\end{equation}

\section{Selected local feature detectors}\label{sec:detectors}

\paragraph{Reference detectors.} Due to large number of existing detectors, we select a sample representative of the breadth of possible approaches. Furthermore, we restrict our attention to detectors that associate a detection strength $\tau$ to each feature (possibly after modifying the implementation of the detector to expose such a value), as needed for selection in the evaluation protocol. That is why we exclude MSER~\citep{matas02local}\footnote{We have experimented using region stability as a detection score surrogate, as defined in~VLFeat~\citep{vedaldi10vlfeat}, but we did not obtain any consistent results.} and Edge Based Regions~\citep{tuytelaars2004matching}.

The selected detectors are listed in~\Cref{tab:detectors}. Detectors are suffixed with -T, -S, -A to emphasise their theoretical viewpoint invariance class (translation, translation+scale and affine respectively). We test a number of detectors representative of traditional techniques such as Harris/Laplace/Hessian cornerness/scale selection and affine adaptation (DoG-S --- aka SIFT-S, SURF-S, Hes-A). We also test FAST-T and BRISK-S, which uses learning to accelerate a standard corner detector. Finally, we test several last-generation detectors that use deep learning: TILDE-T, TCDET-S, LIFT-S, \detnet-T, \detnet-S. \detnet-S is a version of \detnet~\cite{Lenc16} which is evaluated on scaled images, similarly as TCDET-S~\cite{zhang2017learning}. The table also reports their evaluation speed, as this is often a key parameter in applications. Unfortunately, for more recent works~\cite{Savinov2016,DeTone_2018_CVPR_Workshops,Ono2018}, the source code was not available at the time of publication.

\paragraph{Random baseline detectors.} Detectors are also contrasted against a baseline obtained by sampling $n$ features at random~\citep{Verdie15}. We consider: random points (RAND-T), circles (RAND-S) and ellipses (RAND-A). Given a scale $s$ and a $H\times W$ image, the feature center $(u,v)$ is obtained by sampling uniformly at random the set $[s,W-s]\times[s,H-s]$. The scale is sampled as $s \sim \min \{ \|\mathcal{N}(s_{min},(s_{max} - s_{min})^2/4)\|, s_{max} \}$ where $s_{min} = 0.1$ and $s_{max} = 50$ are the minimum and maximum scales. The normal distribution captures the fact that, for most detectors, less features are detected at larger scales. Finally, ellipses are generated by sampling the affine transformation
$
A = \bigl(\begin{smallmatrix}
\cos(\theta) & -\sin(\theta) \\ \sin(\theta) & \cos(\theta)
\end{smallmatrix}\bigr)
\cdot 
\bigl(\begin{smallmatrix}
s \cdot 2^{-a/2} & 0 \\ 0 & s \cdot 2^{a/2}
\end{smallmatrix}\bigr)
$
where $\theta \sim \mathcal{U}(-\pi, \pi)$ and $a \sim \mathcal{U}(0, 2)$ (note that $\sqrt{\det A} = s$ can still be interpreted as scale).

\begin{table}[t]
	\caption{Tested local feature detectors and their speed (in seconds) on four test images from \datasetdet (CPU: single thread Intel Xeon E5-2650 v4; GPU: NVIDIA Tesla M40).}\label{tab:detectors}
	\centering
	\scriptsize
	\definecolor{clr-RAND-T}{rgb}{0.750,0.750,0.750}
\definecolor{clr-RAND-S}{rgb}{0.719,0.750,0.781}
\definecolor{clr-RAND-A}{rgb}{0.592,0.654,0.654}
\definecolor{clr-FAST-T}{rgb}{0.990,0.750,0.723}
\definecolor{clr-SURF-S}{rgb}{0.900,0.680,0.680}
\definecolor{clr-BRISK-S}{rgb}{1.000,0.705,0.852}
\definecolor{clr-DoG-S}{rgb}{0.500,0.750,0.500}
\definecolor{clr-Hes-A}{rgb}{0.500,0.750,0.750}
\definecolor{clr-TILDE-T}{rgb}{0.799,0.699,0.898}
\definecolor{clr-TCDET-S}{rgb}{0.932,0.812,0.932}
\definecolor{clr-LIFT-S}{rgb}{0.750,0.500,0.750}
\definecolor{clr-detnet-T}{rgb}{0.707,0.676,0.900}
\definecolor{clr-detnet-S}{rgb}{0.641,0.619,0.771}
\begin{tabular}{| l | c | c c | c c | c c | c c | c |}
\hline \multicolumn{2}{|c|}{Sequence} & \multicolumn{2}{c|}{\textsc{ajuntament}} & \multicolumn{2}{c|}{\textsc{melon}} & \multicolumn{2}{c|}{\textsc{war}} & \multicolumn{2}{c|}{\textsc{construction}} \\

\multicolumn{2}{|c|}{\# Pixels} & \multicolumn{2}{c|}{0.3M} & \multicolumn{2}{c|}{0.5M} & \multicolumn{2}{c|}{1.0M} & \multicolumn{2}{c|}{3.1M} \\ \hline 

Detector & Impl. & CPU & { \color{blue!80!black} GPU}& CPU & { \color{blue!80!black} GPU}& CPU & { \color{blue!80!black} GPU}& CPU & { \color{blue!80!black} GPU}\\ \hline 
 \cellcolor{clr-FAST-T} FAST-T \citfast & MATLAB & $ 0.10$ & - & $ 0.01$ & - & $ 0.01$ & - & $ 0.03$ & - \\
  \cellcolor{clr-SURF-S} SURF-S  \citsurf & MATLAB & $ 0.14$ & - & $ 0.12$ & - & $ 0.42$ & - & $ 1.20$ & - \\
  \cellcolor{clr-BRISK-S} BRISK-S  \citbrisk & MATLAB & $ 0.26$ & - & $ 0.26$ & - & $ 0.37$ & - & $ 0.52$ & - \\
  \cellcolor{clr-DoG-S} DoG-S  \citdog & VLFeat \cite{vedaldi10vlfeat} & $ 0.14$ & - & $ 0.17$ & - & $ 0.53$ & - & $ 2.80$ & - \\
  \cellcolor{clr-Hes-A} Hes-A  \cithessian & VLFeat \cite{vedaldi10vlfeat} & $ 0.55$ & - & $ 0.56$ & - & $ 2.67$ & - & $ 11.91$ & - \\
  \cellcolor{clr-TILDE-T} TILDE-T  \cittilde & \cittilde & $ 3.42$ & - & $ 5.42$ & - & $ 9.38$ & - & $ 34.99$ & - \\
  \cellcolor{clr-LIFT-S} LIFT-S  \citlift & \citlift & - & { \color{blue!80!black} $ 149.94$ }  & - & { \color{blue!80!black} $ 155.41$ }  & - & { \color{blue!80!black} $ 163.28$ }  & - & { \color{blue!80!black} $ 223.52$ }  \\
  \cellcolor{clr-detnet-T} \detnet-T \citdepdet & \citdepdet & $ 7.64$ & { \color{blue!80!black} $ 0.13$ }  & $ 12.22$ & { \color{blue!80!black} $ 0.18$ }  & $ 25.01$ & { \color{blue!80!black} $ 0.35$ }  & $ 83.92$ & { \color{blue!80!black} $ 1.24$ }  \\
  \cellcolor{clr-detnet-S} \detnet-S \citdepdet & \citdepdet & $ 15.24$ & { \color{blue!80!black} $ 0.35$ }  & $ 24.06$ & { \color{blue!80!black} $ 0.49$ }  & $ 49.64$ & { \color{blue!80!black} $ 0.90$ }  & $ 465.42$ & { \color{blue!80!black} $ 3.00$ }  \\
  \cellcolor{clr-TCDET-S} TCDET-S  \cittcdet & \cittcdet & $ 3.67$ & { \color{blue!80!black} $ 1.42$ }  & $ 7.04$ & { \color{blue!80!black} $ 1.67$ }  & $ 13.24$ & { \color{blue!80!black} $ 2.25$ }  & $ 40.47$ & { \color{blue!80!black} $ 5.35$ }  \\
 \hline
\end{tabular}

\end{table}

\section{Experiments}\label{sec:exps}

\begin{figure}
%    \mbox{}\hspace{-0.3em}
    \includegraphics[scale=0.9]{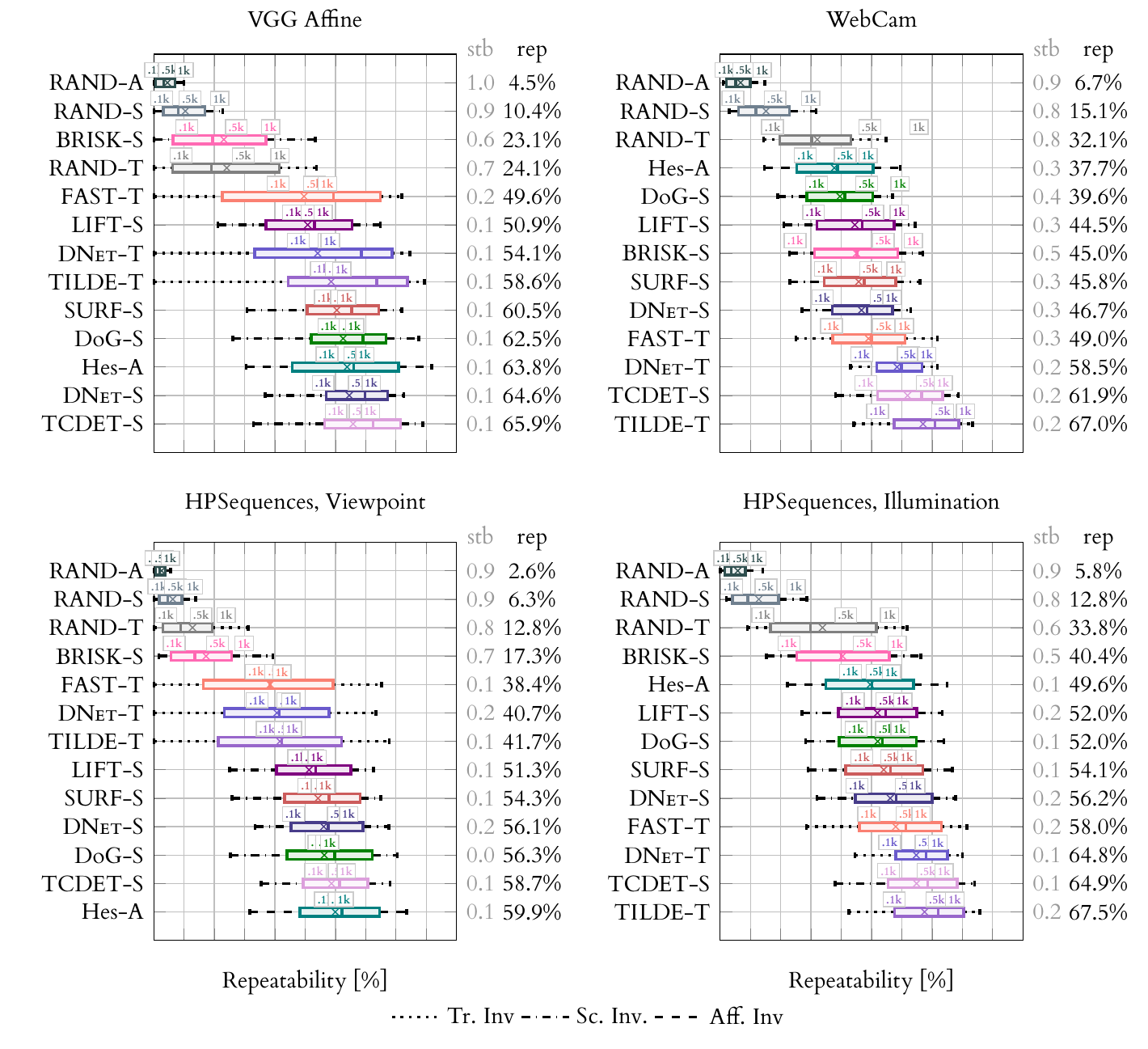}
	%\mbox{}\hspace{-0.5em}
	%\includegraphics[scale=0.9,trim=0 3.1em 0 0,clip]{texfigures/repeval_nfeats_hp.pdf}\\
	\vspace{-2.5em}
\caption{Repeatability of selected detectors on VGG Affine, Webcam, and \datasetdet viewpoint/illumination sequences (HP-V vs HP-I). See \cref{s:det_analysis} for the notation.}\label{fig:rep-hpatches}
\end{figure}

\begin{table}
	\caption{Complete results - average repeatability of the selected detectors on all presented homography datasets.}\label{tab:det_rank}
	\centering
	\setlength{\tabcolsep}{0.10em}
	\scriptsize
	\definecolor{clr-RAND-T}{rgb}{0.750,0.750,0.750}
\definecolor{clr-RAND-S}{rgb}{0.719,0.750,0.781}
\definecolor{clr-RAND-A}{rgb}{0.592,0.654,0.654}
\definecolor{clr-FAST-T}{rgb}{0.990,0.750,0.723}
\definecolor{clr-SURF-S}{rgb}{0.900,0.680,0.680}
\definecolor{clr-BRISK-S}{rgb}{1.000,0.705,0.852}
\definecolor{clr-DoG-S}{rgb}{0.500,0.750,0.500}
\definecolor{clr-Hes-A}{rgb}{0.500,0.750,0.750}
\definecolor{clr-TILDE-T}{rgb}{0.799,0.699,0.898}
\definecolor{clr-TCDET-S}{rgb}{0.932,0.812,0.932}
\definecolor{clr-LIFT-S}{rgb}{0.750,0.500,0.750}
\definecolor{clr-detnet-T}{rgb}{0.707,0.676,0.900}
\definecolor{clr-detnet-S}{rgb}{0.641,0.619,0.771}
\definecolor{gold}{rgb}{0.72, 0.53, 0.04}\definecolor{bronze}{rgb}{0.6, 0.51, 0.48}\definecolor{silver}{rgb}{0.63, 0.47, 0.35}\begin{tabular}{| l | c c c | c c c | c c c | c c c | c c c | c c c | c c c | c |}
\hline \multirow{2}{*}{Det} & \multicolumn{3}{c|}{\textsc{HP-I} \cite{Balntas17}} & \multicolumn{3}{c|}{\textsc{HP-V} \cite{Balntas17}} & \multicolumn{3}{c|}{\textsc{HP-I+V} \cite{Balntas17}} & \multicolumn{3}{c|}{\textsc{VGG} \cite{Mikolajczyk05}} & \multicolumn{3}{c|}{\textsc{WebC} \cite{Verdie15}} & \multicolumn{3}{c|}{\textsc{EFoci} \cite{zitnick2011edge}} & \multicolumn{3}{c|}{\textsc{Hann} \cite{cordes2013high}}  &Avg. $rnk$ \\
& {\color{white!60!black} $stb$} & $rep$ & $rnk$ & {\color{white!60!black} $stb$} & $rep$ & $rnk$ & {\color{white!60!black} $stb$} & $rep$ & $rnk$ & {\color{white!60!black} $stb$} & $rep$ & $rnk$ & {\color{white!60!black} $stb$} & $rep$ & $rnk$ & {\color{white!60!black} $stb$} & $rep$ & $rnk$ & {\color{white!60!black} $stb$} & $rep$ & $rnk$  & \\ \hline 
\cellcolor{clr-TCDET-S} TCDET-S  & {\color{white!60!black} 0.1 } & {\bf \color{silver} 64.91} & {\bf \color{silver} 2} & {\color{white!60!black} 0.1 } & {\bf \color{silver} 58.71} & {\bf \color{silver} 2}  & {\color{white!60!black} 0.1 } & {\bf \color{gold} 61.76} & {\bf \color{gold} 1}  & {\color{white!60!black} 0.1 } & {\bf \color{gold} 65.85} & {\bf \color{gold} 1} & {\color{white!60!black} 0.2 } & {\bf \color{silver} 61.92} & {\bf \color{silver} 2}  & {\color{white!60!black} 0.2 } & {\bf \color{gold} 58.25} & {\bf \color{gold} 1}  & {\color{white!60!black} 0.1 } &  51.55 & {5} &  2.00  \\
 \cellcolor{clr-detnet-S} \detnet-S   & {\color{white!60!black} 0.2 } &  56.20 & {5}  & {\color{white!60!black} 0.2 } &  56.12 & {4} & {\color{white!60!black} 0.2 } & {\bf \color{silver} 56.16} & {\bf \color{silver} 2} & {\color{white!60!black} 0.1 } & {\bf \color{silver} 64.56} & {\bf \color{silver} 2}  & {\color{white!60!black} 0.3 } &  46.70 & {5} & {\color{white!60!black} 0.2 } & {\bf \color{silver} 53.05} & {\bf \color{silver} 2} & {\color{white!60!black} 0.1 } & {\bf \color{silver} 54.72} & {\bf \color{silver} 2} &  3.14  \\
 \cellcolor{clr-TILDE-T} TILDE-T   & {\color{white!60!black} 0.2 } & {\bf \color{gold} 67.52} & {\bf \color{gold} 1}  & {\color{white!60!black} 0.1 } &  41.67 & {7}  & {\color{white!60!black} 0.1 } &  54.37 & {4}  & {\color{white!60!black} 0.1 } &  58.58 & {6}  & {\color{white!60!black} 0.2 } & {\bf \color{gold} 67.03} & {\bf \color{gold} 1}  & {\color{white!60!black} 0.2 } &  50.53 & {4}  & {\color{white!60!black} 0.1 } &  40.37 & {8} &  4.43  \\
 \cellcolor{clr-Hes-A} Hes-A   & {\color{white!60!black} 0.1 } &  49.60 & {9}  & {\color{white!60!black} 0.1 } & {\bf \color{gold} 59.94} & {\bf \color{gold} 1} & {\color{white!60!black} 0.1 } & {\bf \color{bronze} 54.86} & {\bf \color{bronze} 3} & {\color{white!60!black} 0.1 } & {\bf \color{bronze} 63.84} & {\bf \color{bronze} 3}  & {\color{white!60!black} 0.3 } &  37.72 & {10}  & {\color{white!60!black} 0.2 } &  47.30 & {6}  & {\color{white!60!black} 0.0 } & {\bf \color{gold} 58.73} & {\bf \color{gold} 1} &  4.71  \\
 \cellcolor{clr-DoG-S} DoG-S   & {\color{white!60!black} 0.1 } &  52.04 & {7} & {\color{white!60!black} 0.0 } & {\bf \color{bronze} 56.29} & {\bf \color{bronze} 3}  & {\color{white!60!black} 0.1 } &  54.20 & {5}  & {\color{white!60!black} 0.1 } &  62.53 & {4}  & {\color{white!60!black} 0.4 } &  39.56 & {9} & {\color{white!60!black} 0.2 } & {\bf \color{bronze} 51.12} & {\bf \color{bronze} 3} & {\color{white!60!black} 0.0 } & {\bf \color{bronze} 54.71} & {\bf \color{bronze} 3} &  4.86  \\
 \cellcolor{clr-SURF-S} SURF-S   & {\color{white!60!black} 0.1 } &  54.05 & {6}  & {\color{white!60!black} 0.1 } &  54.25 & {5}  & {\color{white!60!black} 0.1 } &  54.16 & {6}  & {\color{white!60!black} 0.1 } &  60.45 & {5}  & {\color{white!60!black} 0.3 } &  45.76 & {6}  & {\color{white!60!black} 0.2 } &  49.10 & {5}  & {\color{white!60!black} 0.0 } &  51.76 & {4} &  5.29  \\
 \cellcolor{clr-detnet-T} \detnet-T  & {\color{white!60!black} 0.1 } & {\bf \color{bronze} 64.79} & {\bf \color{bronze} 3}  & {\color{white!60!black} 0.2 } &  40.65 & {8}  & {\color{white!60!black} 0.1 } &  52.51 & {7}  & {\color{white!60!black} 0.1 } &  54.15 & {7} & {\color{white!60!black} 0.2 } & {\bf \color{bronze} 58.47} & {\bf \color{bronze} 3}  & {\color{white!60!black} 0.2 } &  46.84 & {7}  & {\color{white!60!black} 0.1 } &  40.82 & {7} &  6.00  \\
 \cellcolor{clr-LIFT-S} LIFT-S   & {\color{white!60!black} 0.2 } &  51.96 & {8}  & {\color{white!60!black} 0.1 } &  51.27 & {6}  & {\color{white!60!black} 0.1 } &  51.61 & {8}  & {\color{white!60!black} 0.1 } &  50.85 & {8}  & {\color{white!60!black} 0.3 } &  44.48 & {8}  & {\color{white!60!black} 0.2 } &  45.84 & {8}  & {\color{white!60!black} 0.1 } &  41.20 & {6} &  7.43  \\
 \cellcolor{clr-FAST-T} FAST-T   & {\color{white!60!black} 0.2 } &  57.99 & {4}  & {\color{white!60!black} 0.1 } &  38.43 & {9}  & {\color{white!60!black} 0.1 } &  48.04 & {9}  & {\color{white!60!black} 0.2 } &  49.63 & {9}  & {\color{white!60!black} 0.3 } &  49.01 & {4}  & {\color{white!60!black} 0.2 } &  39.74 & {9}  & {\color{white!60!black} 0.1 } &  39.44 & {9} &  7.57  \\
 \cellcolor{clr-BRISK-S} BRISK-S   & {\color{white!60!black} 0.5 } &  40.43 & {10}  & {\color{white!60!black} 0.7 } &  17.25 & {10}  & {\color{white!60!black} 0.5 } &  28.64 & {10}  & {\color{white!60!black} 0.6 } &  23.14 & {11}  & {\color{white!60!black} 0.5 } &  45.03 & {7}  & {\color{white!60!black} 0.6 } &  29.11 & {10}  & {\color{white!60!black} 0.7 } &  12.58 & {10} &  9.71  \\
 \cellcolor{clr-RAND-T} RAND-T   & {\color{white!60!black} 0.6 } &  33.81 & {11}  & {\color{white!60!black} 0.8 } &  12.78 & {11}  & {\color{white!60!black} 0.7 } &  23.11 & {11}  & {\color{white!60!black} 0.7 } &  24.11 & {10}  & {\color{white!60!black} 0.8 } &  32.10 & {11}  & {\color{white!60!black} 0.6 } &  28.09 & {11}  & {\color{white!60!black} 0.8 } &  12.03 & {11} &  10.86  \\
 \cellcolor{clr-RAND-S} RAND-S   & {\color{white!60!black} 0.8 } &  12.78 & {12}  & {\color{white!60!black} 0.9 } &  6.27 & {12}  & {\color{white!60!black} 0.9 } &  9.47 & {12}  & {\color{white!60!black} 0.9 } &  10.41 & {12}  & {\color{white!60!black} 0.8 } &  15.14 & {12}  & {\color{white!60!black} 0.8 } &  17.11 & {12}  & {\color{white!60!black} 0.9 } &  5.61 & {12} &  12.00  \\
 \cellcolor{clr-RAND-A} RAND-A   & {\color{white!60!black} 0.9 } &  5.82 & {13}  & {\color{white!60!black} 0.9 } &  2.59 & {13}  & {\color{white!60!black} 0.9 } &  4.17 & {13}  & {\color{white!60!black} 1.0 } &  4.50 & {13}  & {\color{white!60!black} 0.9 } &  6.74 & {13}  & {\color{white!60!black} 0.9 } &  8.09 & {13}  & {\color{white!60!black} 0.8 } &  2.61 & {13} &  13.00  \\
 \hline
\end{tabular}

\end{table}

\paragraph{Datasets.} While we use several datasets in our evaluation (\cref{tab:datasets-det}), we mainly focus on \datasetdet which builds on the images of \dataset~\cite{Balntas17}. This contains image sequences in a similar format to the original VGG Affine dataset, but with an order of magnitude more sequences, divided in viewpoint (HP-V) and illumination (HP-I) changes.

\paragraph{Aggregated evaluation.} We first evaluate the average repeatability of the detectors (\cref{fig:rep-hpatches,tab:det_rank}) as defined in~\cref{e:rep}. For the older \textbf{VGG Affine}, translation invariant detectors such as TILDE-T are competitive in median/average with more invariant detectors (-S, -A), but in 10\% of the cases fail catastrophically (see the whiskers). The latter problem is solved by scale invariance and the best detectors use Hessian or Laplacian-based scale selection (-S). On the \textbf{Webcam} dataset, which contains only illumination changes, RAND-T is surprisingly competitive (mostly due to the fact that scale is always selected consistently), on part with more complex -S and -A detectors. TILDE-S, which is learned on this dataset, is unsurprisingly the winner.

Next, we look at \textbf{\datasetdet}, starting from the viewpoint sequences (HP-V). Compared to the previous datasets, the RAND-T,S,A baselines perform much worse, confirming that this data is significantly harder. The best detectors are variants of the Hessian one, which is popular in instance retrieval~\citep{arandjelovic2012three, perd2009efficient}, and scale selection (-S) brings in general an advantage; however, the benefits of Baumberg \cite{baumberg00reliable} affine adaptation (-A) is small. In general, the top six detectors perform similarly. For the illumination sequences (HP-I, Webcam), since scale does not change, -T detectors are advantaged. The best performance is again achieved with the TILDE-T, which therefore generalises beyond the Webcam dataset.

From the relatively high performance of the RAND-T detector, we can see that  it is crucial to compare detectors of similar classes. This also justifies the use of ellipse overlap over the simpler distance of keypoint centres for detector evaluation.

For the stability across detection thresholds~\eqref{e:stb}, we see in~\cref{tab:det_rank} that the majority of the best performing detectors have their stability errors under $10\%$. However, the stability is much lower for the BRISK and RANDOM detectors, which indicates that the BRISK detection scores are not predictive of the detector performance.

%\todo{CHECK} Again, for the illumination changes, most of the detectors are more sensitive to the detection threshold compared to the viewpoint tasks, however still much less than the random detector or the BRISK detector, which does not seem to be performing well on any task.

\begin{figure}
	\hspace{-2.2em}
	\includegraphics[width=1.05\linewidth]{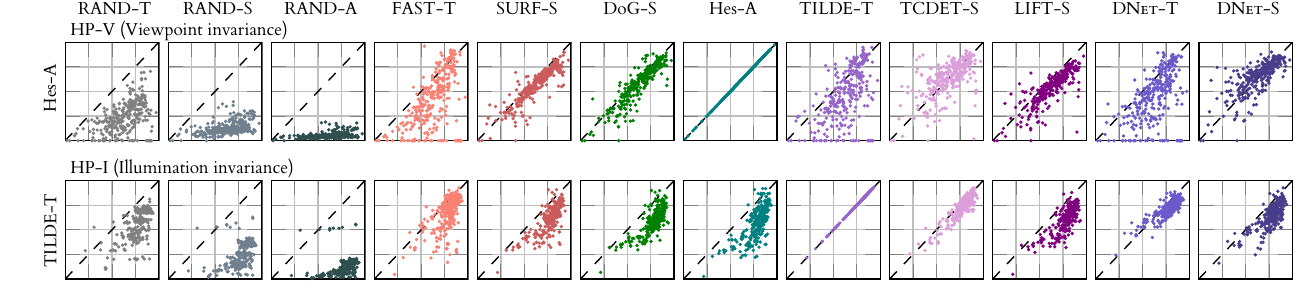}%
	\vspace{-1.em}
	\caption{Comparison of repeatability distributions of pair of detectors on different subsets \datasetdet (HP-V and HP-I), $n=1000$. The x-axis is a repeatability of the reference detector (specified by row) and y-axis is repeatability of the selected detector, specified by column. Each point in a plot represents a repeatability of a single image pair.} \label{fig:rep-crossdist}
\end{figure}

\begin{figure}
	%\vspace{-1.2em}
	\includegraphics[width=1.\linewidth]{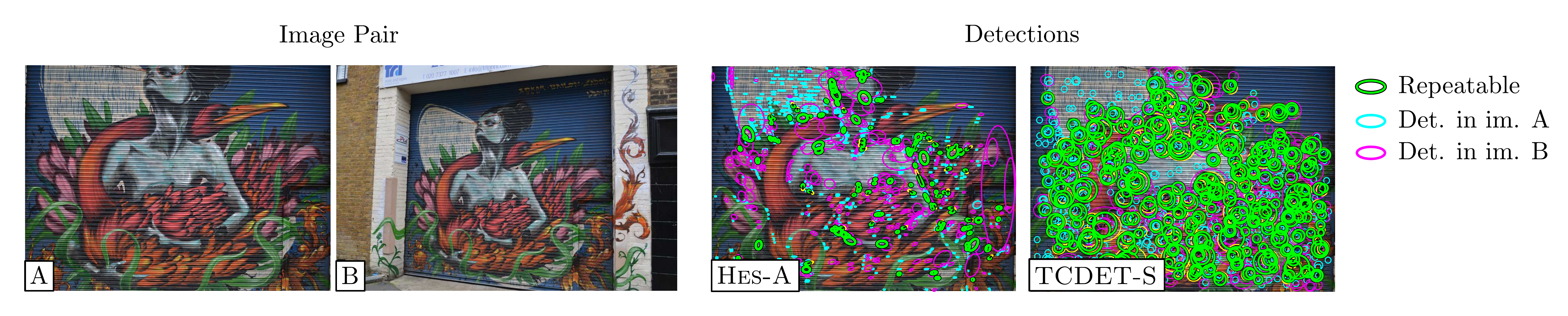}%
%	\vspace{-1.5em}
	\caption{Example of an image pair where a trained detector ($rep(\textsc{TCDET-S}) =69.8$) achieves better performance compared to a traditional detector ($rep(\textsc{Hes-A})=28.57$).}\label{fig:det-example}
\end{figure}

\paragraph{Non-modal performance.} A limitation of the analysis above is that it relies on aggregated measures that may hide particular example cases where a given detector has a significant advantage, such as an extreme viewpoint change. To analyse this possibility, in~\cref{fig:rep-crossdist} we plot the repeatability of each detector (y-axis) against the one of the best reference detector (Hes-A and TILDE-T) for all images in the viewpoint (HP-V) and illumination (HP-I) sequences. Points above the diagonal mean that the tested detector (column) obtained higher repeatability on a specific image pair than the reference detector (row). Please note that the distribution of the visualised points across y-axis would give us \cref{fig:rep-hpatches}. We can se that TILDE-T tends to uniformly dominate other detectors in HP-I, but for HP-V the best overall detector Hes-A is occasionally outperformed by other detectors such as \detnet or TCDET-S, which can therefore be complementary (qualitative example in \cref{fig:det-example}).

Finally, in~\cref{tab:det_rank} we test the consistency of the results across several more datasets (HP-I, -V, -I+V, VGG, Webcam, Edge Foci, Hannover), reporting repeatability, stability and the rank of each detector together with its average rank. Remarkably, detectors learned using the covariance constraint with scale invariance lead the performance across the selected datasets. However traditional detectors generally outperform the trained detectors on tasks where a viewpoint invariance is important. Nonetheless, for learnt detectors this might be mitigated with additional data augmentation or training on datasets with more viewpoint variations.
Similarly, the random detectors set a baseline performance for both repeatability and stability across detector's selectivity.

\section{Discussion}

While learning is poised to change local feature detection, developing a new generation of algorithms almost invariably requires the introduction of improved benchmark datasets. Object detection had PASCAL VOC, deep learning had ImageNet, and handcrafted detectors had VGG Affine. In this paper, we have proposed to improve and extend VGG Affine's protocol to large scale evaluation.
While performance of the whole local feature pipeline is important, ability to compare detection performance of different algorithms, without undue influence of the selected description and matching algorithm, is crucial. It not only allows to assess geometric precision of a detector, but in combination with descriptor evaluation it allows to pinpoint the main source of improvement. 
We are hoping that this detailed analysis will catalyse further progress and advance our understanding of machine learning applied to local feature detection.

%Often, the final performance of the whole processing pipeline is important, 
%The benchmark is a natural evolution of VGG Affine and builds on the \dataset and other datasets that have already been road tested for the evaluation of local descriptors.
  
Using this benchmark, we have assessed several traditional and deep detectors. We have showed that, while machine learning clearly helps for illumination invariance, for viewpoint invariance traditional methods are still surprisingly competitive. This suggests that there is still significant potential for progress in this area.

%A well known limitation of detector evaluation is that detector performance might not be fully predictive of the final  (which might be the case of LIFT features which are trained jointly for detection and description). However, it is important to learn which individual component leads to improved performance  
%However it is still important to be able to asses what drives the progress by evaluating performance of individual components.

\subsection*{Acknowledgements}
This research was supported by ERC 677195-IDIU and Programme Grant Seebibyte \\ EP/M013774/1.
We would like to thank Dmytro Mishkin for useful feedback on the first versions of this manuscript and Aravindh Mahendran for proof reading.
Additionally, we are thankful for the constructive feedback of the BMVC reviewers which helped to improve this work.
 
\bibliography{literature}
%\newpage\input{scratch}

\end{document}